# APTOS-2024 challenge report: Generation of synthetic 3D OCT images from fundus photographs

**Authors:**

Bowen Liu[1,a], Weiyi Zhang[1,a], Peranut Chotcomwongse[2,a], Xiaolan Chen[1,a], Ruoyu Chen[1,a], Pawin Pakaymaskul[2,a], Niracha Arjkongharn[2,a], Nattaporn Vongsa[2,a], Xuelian Cheng[3,a], Zongyuan Ge[3,a], Kun Huang[4,b], Xiaohui Li[4,b], Yiru Duan[4,b], Zhenbang Wang[4,b], BaoYe Xie[4,b], Qiang Chen[4,b], Huazhu Fu[5,b], Michael A. Mahr[6,b], Jiaqi Qu[7,b], Wangyiyang Chen[8,b], Shiye Wang[9,b], Yubo Tan[10,b], Yongjie Li[10,b], Mingguang He[1,11,12,a], Danli Shi[1,11, a], Paisan Ruamviboonsuk[2,a]

**Affiliation:**

1. School of Optometry, The Hong Kong Polytechnic University, Kowloon, Hong Kong

2. Department of Ophthalmology, College of Medicine, Rangsit University, Rajavithi Hospital, Thailand

3. eResearch Center and Faculty of Engineering, Monash University, Clayton, VIC, 3800 Australia

4. Nanjing University of Science and Technology, Nanjing 210000, Jiangsu, China

5. Institute of High Performance Computing (IHPC), Agency for Science, Technology and Research (A*STAR), Singapore

6. Mayo Clinic Department of Ophthalmology, Rochester, MN USA

7. China Medical University, Shenyang 110122, Liaoning, China

8. Northeastern University, Shenyang 110819, Liaoning, China

9. Henan Kaifeng College of Science Technology and Communication, Kaifeng 475001, China

10. University of Electronic Science and Technology of China, Chengdu 610054, China

11. Research Centre for SHARP Vision (RCSV), The Hong Kong Polytechnic University, Kowloon, Hong Kong

12. Centre for Eye and Vision Research (CEVR), 17W Hong Kong Science Park, Kowloon, Hong Kong

**Corresponding to:**

Prof. Paisan Ruamviboonsuk, Department of Ophthalmology, College of Medicine, Rangsit University, Rajavithi Hospital. Email: paisan.trs@gmail.com

Dr. Danli Shi, Research Assistant Professor, School of Optometry, The Hong Kong Polytechnic University, Kowloon, Hong Kong. Email: danli.shi@polyu.edu.hk

Prof. Mingguang He, Chair Professor of Experimental Ophthalmology, School of Optometry, The Hong Kong Polytechnic University, Kowloon, Hong Kong. Email: mingguang.he@polyu.edu.hk

[a] Challenge organizer. [b] Contributed results of their algorithms presented in the paper.

**Abstract**

Optical Coherence Tomography (OCT) provides high-resolution, three-dimentional (3D), and non-invasive visualization of retinal layers in vivo, serving as a critical tool for lesion localization and disease diagnosis. However, its widespread adoption is limited by equipment costs and the need for specialized operators. In comparison, two-dimensional color fundus photography offers faster acquisition and greater accessibility with less dependence on expensive devices. Although generative artificial intelligence (AI) has demonstrated promising results in medical image synthesis, translating 2D fundus images into 3D OCT images presents unique challenges due to inherent differences in data dimensionality and biological information between modalities. To advance generative models in the fundus-to-3D-OCT setting, the Asia Pacific Tele-Ophthalmology Society (APTOS-2024) organized a challenge titled "Artificial Intelligence-based OCT Generation from Fundus Images". This paper details the challenge framework (referred to as APTOS-2024 Challenge), including: (1) the benchmark dataset, (2) evaluation methodology featuring two fidelity metrics—image-based distance (pixel-level OCT B-scan similarity) and video-based distance (semantic-level volumetric consistency), and (3) analysis of top-performing solutions. The challenge attracted 342 participating teams, with 42 preliminary submissions and 9 finalists. Leading methodologies incorporated innovations in hybrid data preprocessing / augmentation (cross-modality collaborative paradigms), pre-training on external ophthalmic imaging datasets, integration of vision foundation models, and model architecture improvement. The APTOS-2024 Challenge is the first benchmark demonstrating the feasibility of fundus-to-3D-OCT synthesis as a potential solution for improving ophthalmic care accessibility in under-resourced healthcare settings, while helping to expedite medical research and clinical applications.

## 1. Introduction

Optical coherence tomography (OCT) has become an essential ophthalmic imaging technique, providing high-resolution cross-sectional and volumetric visualization of multi-layered retinal structures. Its ability to detect pathological changes in specific retinal layers, such as macular edema[1] and choroidal neovascularization, has made it invaluable for clinical diagnosis[2]. However, OCT adoption faces limitations due to the substantial equipment costs, technical expertise required for operation, and restricted availability in under-resourced healthcare settings[3,4]. In comparison, fundus photography is a more accessible and cost-effective imaging tool but provides two-dimensional (2D) en-face views of the retina. Fundus images alone fail to deliver the detailed information necessary for precise localization and monitoring of chorioretinal abnormalities that OCT provides.

The ability to synthesize 3D OCT data from 2D fundus images offers a promising solution that balances accessibility with diagnostic precision in ophthalmic practice. By converting widely accessible 2D fundus images into detailed 3D layered anatomy,[5] this technology could facilitate critical early detection of retinal and choroidal diseases while reducing dependency on costly OCT equipment. It could also enable large-scale reevaluation of historical fundus image collections. The implementation of this cross-modal translation, however, confronts substantial challenges rooted in fundamental differences between imaging modalities. Fundus photography and OCT capture distinct aspects of retinal structure - while the former captures surface reflectance, the latter encodes depth-resolved tissue reflectance and scattering properties. Successfully bridging these modalities requires not only reconstructing unobserved depth information from 2D projections but also ensuring anatomical plausibility across generated volumetric data and preserving clinically relevant pathological features throughout the translation process. These requirements present unique obstacles that existing medical image synthesis approaches have yet to fully overcome. Most current work focuses on 2D OCT image generation[6-9], with the majority relying on Generative Adversarial Networks (GANs)[10]. These methods generate synthetic images from noise through a generator, and often require training different models for various data categories. There is a notable lack of research focused on generating 3D OCT images.

To advance generative models in the fundus-to-3D-OCT setting, we initiated the APTOS-2024 challenge, which focuses on generating 3D OCT images from fundus photographs. The benchmark dataset comprises 1,247 fundus photographs paired with high-resolution OCT images from 693 patients. The challenge comprised two main phases. In the preliminary phase, teams utilized the provided training dataset for algorithm development, with the option to refine their approaches based on performance evaluation using a preliminary validation dataset. The final phase was open to all participating teams, with winners ultimately determined through objective assessment on an independent final test dataset. The objective assessment features two fidelity metrics—image-based distance (i.e., the pixel-level similarity between each B-scan of the real and generated OCT in a one-to-one comparison) and video-based distance (i.e., highlighting the semantic-level similarity between the real and generated OCT data).

This report on the challenge presents distinct insights. Below are the main points that emphasize its contributions:

1. Benchmark resource: It establishes a standardized benchmark for fundus-to-3D-OCT translation, comprising a multimodal retinal image dataset with an evaluation protocol on generation fidelity.

2. Methodological Insights: The systematic analysis of top-performing methodologies reveals critical insights into cross-modal medical image generation, particularly the effectiveness of the collaborative technical efforts for bridging the fundus-to-3D-OCT gap.

3. Clinical Relevance: The challenge outcomes demonstrate this technology's potential to transform ophthalmic diagnostics and provide critical references for future work on generating more detailed 3D retinal structures from fundus images or other possible modalities.

The rest of this paper is organized as follows: Section 2 outlines the challenge setup, dataset, and evaluation protocol. Section 3 gives the summary of top-performing methods. Section 4 reports and discusses the challenge results. Section 5 examines the characteristics, significance, limitations, and future directions of the challenge. Finally, Section 6 concludes the paper.

**2. Material and methods**

2.1 Challenge organization

APTOS-2024 challenge was organized by the Asia Pacific Tele-Ophthalmology Society (APTOS), co-hosted by Rajavithi Hospital in Thailand, sponsored and recognized by the Medical Services Department of the Thai Ministry of Public Health, and receives exclusive technical support from Alibaba Cloud Tianchi platform. The challenge aimed at providing a benchmark for developing and evaluating the algorithms that are used for 3D OCT image generation from fundus photographs. It is a one-time event with a fixed submission deadline.

The APTOS-2024 challenge was composed of two main phases: the preliminary phase and the final phase. Fig. 1a shows the overall challenge workflow. In the preliminary phase, participating teams can access the complete training dataset (including fundus images paired with 3D OCT images) and the input of the validation dataset (which contains only fundus images). All teams utilized the provided training dataset for algorithm development, with the option to refine their approaches based on performance evaluation using the preliminary validation dataset. The final phase was open to all participating teams, with winners ultimately determined through objective assessment on an independent final test dataset. The APTOS-2024 challenge is open to everyone. Although members of the APTOS Council and Alibaba Group employees were eligible to participate, their results will not be considered in the ranking process.

Research teams registered and accessed the datasets through the challenge website, they could review rules, submit results, and check rankings (https://tianchi.aliyun.com/competition/entrance/532266/). The system retained only the latest submission for final evaluation, with detailed guidelines provided. The preliminary phase ran from August 16 to November 30, 2024. The final phase began on December 13, 2024, when participants downloaded the final test dataset on the event day. Teams could submit up to 100 entries by 18:00 (UTC+8) on the final day. Any incomplete submissions were considered invalid. The top 5 teams on the leaderboard were subject to a mandatory code review by 18:00 (UTC+8) on December 14, 2024. If any issues such as code plagiarism, cheating, or failure to reproduce the optimal results were identified, the spot in the finals would be given to the next highest-ranking team that passed the review. The top 5 teams on the final leaderboard will be formally recognized and have the opportunity to present their AI models.

2.2 Benchmark dataset

A total of 1,247 3D OCT images paired with fundus images were collected from 693 patients. The data division (patient-level split) is detailed in Table 1. Each 3D OCT data is represented by six different directional OCT B-scans. The released OCT dataset is organized as shown in a frame-wise structure. We used such organization because it makes loading faster for possible sparse training. The fundus images corresponding to the OCT data are stored in JPG format. All paired information was provided in a csv file.

For participants, the training data can include not only the data provided by the challenge but also publicly available data, including (open) pre-trained networks. If private datasets or networks are used for training, they must be made public to ensure fairness.

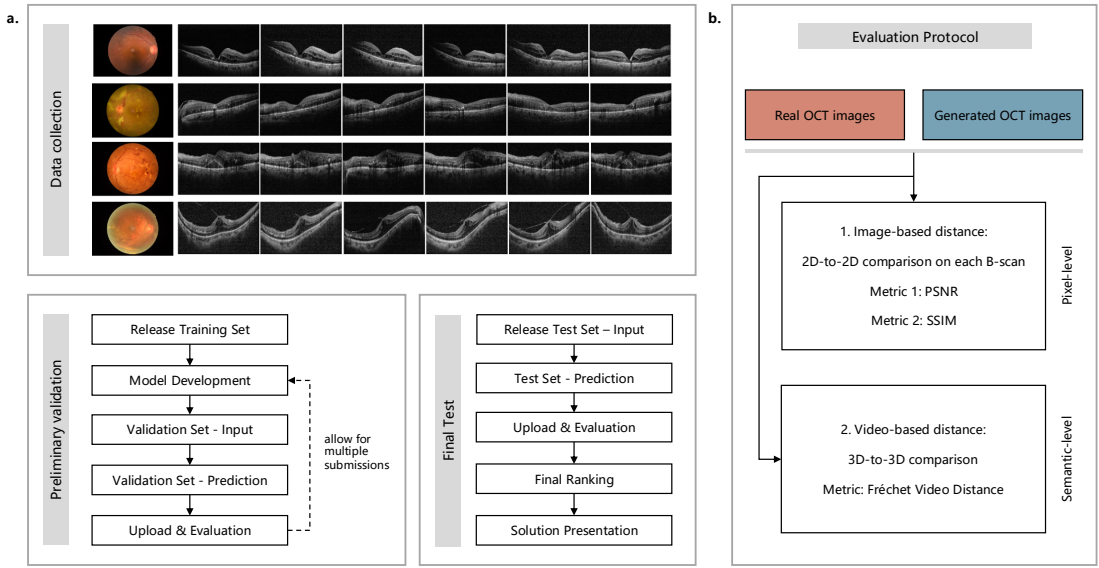

**Fig. 1. The overall challenge workflow with the detailed evaluation protocol.** a) The challenge comprised two main phases. b) Evaluation methodology featuring two fidelity metrics—image-based distance (pixel-level OCT B-scan similarity) and video-based distance (semantic-level volumetric consistency).

**Table 1.** Dataset division for the APTOS-2024 challenge.

| Subset | Pair Num | Patient Num |
| --- | --- | --- |
| Total | 1247 | 693 |
| Training Set | 750 | 393 |
| Preliminary Round Test Set | 254 | 150 |
| Final Round Test Set | 243 | 150 |

2.3 Evaluation methodology

The assessment framework incorporates two complementary fidelity metrics to evaluate the quality of generated OCT volumes: image-based distance and video-based distance. As shown in Fig. 1b, the image-based distance quantifies pixel-level similarity through direct one-to-one comparison of corresponding B-scans between real and synthetic OCT data, while the video-based distance assesses semantic-level consistency across the complete 3D reconstruction, ensuring structural plausibility in the generated 3D output.

For frame-level evaluation, Peak Signal-to-Noise Ratio (PSNR) and Structural Similarity Index Measure (SSIM)[11] were computed on a per-frame basis, with results averaged across six representative frames. Higher PSNR values indicate better reconstruction fidelity, whereas SSIM scores (ranging from -1 to 1) approach 1 for optimal perceptual similarity. At the video level, the Fréchet Video Distance[12] (FVD) measured semantic similarity between predicted and ground-truth 3D images, where lower values reflect stronger alignment (on feature distribution).

Notably, the evaluation prioritized FVD as the primary ranking criterion. Conventional pixel-wise metrics (PSNR/SSIM) were intentionally assigned secondary roles, as their sensitivity to preprocessing artifacts, particularly those introduced by data augmentation, could skew performance interpretation and reduce assessment robustness. Above combination of low-level and high-level reconstruction accuracy metrics provides research teams with comprehensive insights for model optimization priorities.

## 3. Summary of top-ranked algorithms

The top four teams in the competition rankings are NJUST-EYE, MC1, Mickey Mouse Clubhouse, and ViCBiC. The following will introduce the teams using the aliases Algorithm 1-4. In this section, we provide detailed descriptions of the methods employed by the top-ranked algorithms, with a focus on data pre-processing, data augmentation, model architecture, and external knowledge. The details of Algorithm 1-4 are given in Appendix.

**Table 2.** Summary of data pre-processing steps in top-ranked algorithms. Where Fundus-ROI is removing redundant black border in the fundus image; Fundus-S-ROI is selecting specialized regions based on the axial correspondence; OCT-Crop is removing equipment markers in the OCT data; Resize is if both fundus or OCT images are resized before model development.

| Algorithm | Fundus-ROI | Fundus-S-ROI | OCT-Crop | Resize |
|---|---|---|---|---|
| 1 | √ | × | √ | √ |
| 2 | √ | √ | √ | √ |
| 3 | √ | × | √ | √ |
| 4 | √ | √ | √ | √ |

3.1 Data pre-processing

Prior to initiating model development, it is essential to minimize redundant information within the dataset and eliminate samples exhibiting significant distributional discrepancies to ensure training data homogeneity. We analyzed the preprocessing steps adopted by different teams. For fundus images (the input modality for the generative models), all algorithms employed redundant black border removal through Region of Interest (ROI) cropping. Notably, Algorithm 2 and Algorithm 4 implemented specialized ROI selection based on the axial correspondence between each OCT B-scan and their en-face fundus projections. Specifically, Algorithm 2 masked non-ROI regions (zeroing pixels outside target areas), while Algorithm 4 performed direct cropping of the designated ROIs. For OCT data, all algorithms employed cropping operations to remove redundant information (such as device-specific markers from imaging equipment) while maximally preserving complete retinal structural information. A summary of data-preprocessing steps is illustrated in Table 2.

3.2 Data Augmentation

Deep neural networks typically require large amounts of training data to achieve satisfactory performance. When data is limited, data augmentation is commonly employed to expand the dataset, improve model robustness, and prevent overfitting. In the context of image data, current data augmentation techniques primarily involve applying algorithmic transformations to images, such as geometric modifications or noise injection, to increase data diversity. However, unlike common augmentation strategies, cross-modality generation task like this in the challenge also demands cross-modality data augmentation. Modality-independent augmentation methods are significantly constrained in this scenario, as they cannot effectively account for the intrinsic differences between modalities.

Algorithm 1 proposed a simple yet effective cross-modality augmentation method. As shown in Fig. 2, it involves flipping the fundus image and the corresponding OCT images in a collaborative paradigm. For modality-independent augmentation, Algorithm 4, for instance, employed a variety of data augmentation techniques, including random adjustments of contrast, brightness, and saturation, as well as gamma correction, Gaussian blur, and the addition of Gaussian noise, applied to fundus images.

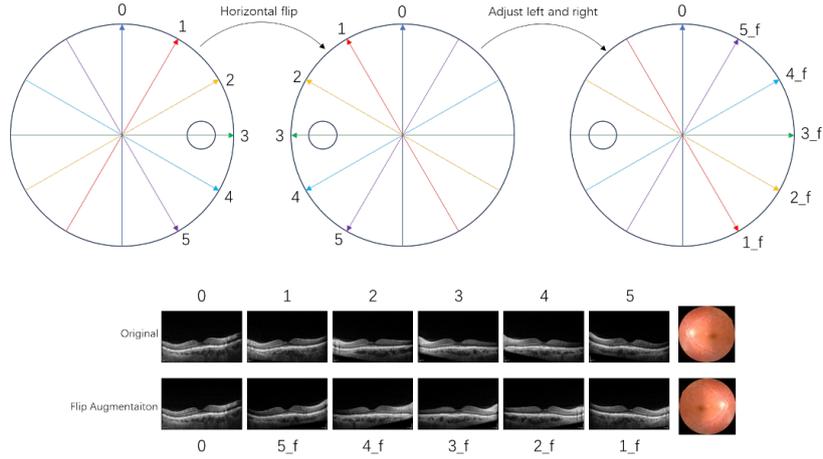

- Color fundus flipped directly horizontally
- OCT_0 unchanged (enhanced frame 0)
- OCT_1 flipped horizontally to become OCT_1_f (enhanced frame 5)
- OCT_2 flipped horizontally to become OCT_2_f (enhanced frame 4)
- OCT_3 flipped horizontally to become OCT_3_f (enhanced frame 3)
- OCT_4 flipped horizontally to become OCT_4_f (enhanced frame 2)
- OCT_5 flipped horizontally to become OCT_5_f (enhanced frame 1)

**Fig. 2. Flip enhancement employed by Algorithm 1.** This process involves flipping the fundus image and the corresponding OCT images horizontally.

3.3 Model architecture

Generative models, a fundamental class of machine learning methods, learn to represent and model the underlying distribution of training data. Ideally, these models should satisfy three critical requirements: (1) high-quality sampling—producing outputs indistinguishable from real data; (2) mode coverage and diversity—faithfully capturing the full variability of complex training datasets; and (3) computationally efficient sampling—enabling the wide adoption for practical real-world applications. However, existing methodologies face inherent limitations in simultaneously meeting all these demands. For instance, while GANs excel in sample quality,[13-16] their training often suffers from instability and mode collapse; Variational Auto-Encoder (VAE) guarantees diversity[17] but may generate blurry or low-fidelity outputs; and diffusion models, despite their robustness,[18] are unfortunately hindered by slow sampling speeds. Therefore, participating teams implemented targeted modifications to their base generative models to enhance their suitability for this challenge task.

Algorithm 1 was built upon the latent diffusion model (LDM)[19] architecture with three key structural innovations. First, the whole framework integrates a fundus encoder for extracting the condition feature from the entire fundus image and a spatial map encoder to preserve cross-slice anatomical relationships, while an image-level encoder projects individual OCT B-scans into a compressed latent space for efficient diffusion training. Second, the denoising U-Net[20] employs a hierarchical attention system: cross-attention layers dynamically fuse fundus-derived semantic guidance with 3D OCT latent representations, while intra-batch self-attention mechanisms model implicit correlations between sequential B-scans by treating each six-slice image batch as a spatiotemporal sequence. Third, the two-stage generation pipeline first synthesizes latent representations through conditioned iterative denoising using residual blocks with integrated attention, then reconstructs high-fidelity volumetric OCT data via the image-level decoder. This architecture uniquely addresses the fundus-to-3D-OCT translation challenge by simultaneously maintaining computational tractability through latent space processing, enforcing anatomical plausibility via cross-modal conditioning, and ensuring consistency through learned inter-slice dependencies.

Algorithm 2 utilizes the Pix2Pix conditional GAN[21], as originally described, using the publicly available implementation at https://github.com/junyanz/pytorch-CycleGAN-and-pix2pix as the foundation for subsequent customizations. The generator is based on a U-Net architecture, while the discriminator employs a 70×70 PatchGAN classifier. The model uses a vanilla GAN binary cross-entropy loss function, as outlined in the original paper.

Algorithm 3 trained six sub-models, each generating one B-scan of the 3D OCT images. Each sub-model learns different information but has the same structure. Below, only the details of the sub-models are described. Each sub-model is based on a noise-injected conditional GAN, Dynamic-Pix2Pix[22]. This method incorporates dynamic neural network theory to enhance the generalization capability of Pix2Pix, especially for small sample datasets. When the input is a fundus image, Dynamic-Pix2Pix learns the direct mapping between the fundus and OCT images. When the input is a noise image, the generator focuses on learning the distribution characteristics of the target OCT image. The generator uses a U-Net architecture with an additional noise bottleneck module, which limits the transmission of noise information. Positioned in the middle of the U-Net, this module reduces the number of noise input channels and applies max-pooling layers to filter out noise, ensuring that the generator focuses on the critical image structures rather than noise.

Algorithm 4 utilizes a model architecture designed for specialized pre-processed fundus images, where the images are extracted along six sequences corresponding to the six OCT scanning directions. The first part of the process involves using three layers of Multilayer Perceptrons (MLP) to process the input sequence, generating a feature map. This output feature map is then passed through a U-Net architecture, which applies a sigmoid activation function at the output layer to generate 3D OCT images. The U-Net consists of five layers of symmetric encoders and decoders, each containing 16 channels per layer. Each layer of the U-Net includes three Conv (3×3)-BatchNorm-ReLU blocks, which help capture hierarchical features and refine the generated images.

Finally, a summary of model architectures is illustrated in Table 3.

Table 3. Summary of model architectures with 3D learning in top-ranked algorithms.

| Algorithm | Base model | Specific module | 3D Learning |
|---|---|---|---|
| 1 | latent diffusion model | self-attention-based | 2D compression → 3D diffusion (2.5D) |
| 2 | Pix2Pix | -- | frame-wise learning (2D) |
| 3 | Dynamic-Pix2Pix | noise bottleneck | 2D sub-models (2D) |
| 4 | auto encoder | -- | direct 3D (3D) |

3.4 External Knowledge

In the development of Algorithm 1, the integration of external knowledge through publicly available pretrained models and additional OCT datasets significantly enhances the model's performance in generating high-quality 3D OCT images.

During the 2D compression stage, a robust VAE from Stable Diffusion is used to extract latent representations of OCT B-scan images (https://huggingface.co/stable-diffusion-v1-5/stable-diffusion-v1-5). In the preparation of 3D diffusion training, Algorithm 1 utilized three publicly available OCT datasets (NEHUT2021[23], OLIVES Dataset[24], and OCTA-500[25]) to train a unconditional latent diffusion model, which significantly refines its capability to generate accurate 3D images. Additionally, in the conditional mechanism, three pretrained foundation models (RETFound[26], MM-Retinal[27], FLAIR[28]) were leveraged to extract retinal features from the whole fundus images. These fundus models contribute to a deeper understanding of retinal structures, which is crucial for producing clinically relevant and detailed 3D OCT outputs.

Through the integration of these external resources, Algorithm 1 underscores the importance of external knowledge in improving model performance, ultimately enabling the generation of high-quality 3D OCT images.

Table 4. The performance of the top four algorithms in the final round, along with their results in the preliminary round and baseline comparisons.

| Algorithm | Final round | | | Preliminary round | | |
|---|---|---|---|---|---|---|
| | FVD↓ | SSIM↑ | PSNR↑ | FVD↓ | SSIM↑ | PSNR↑ |
| 1 | **624.5898** | 0.1048 | 13.6502 | **633.3071** | 0.0989 | 13.6556 |
| 2 | 630.8068 | 0.1433 | **15.1206** | 713.3958 | **0.1651** | **14.9492** |
| 3 | 640.7700 | **0.1498** | 14.5759 | 654.4793 | 0.1582 | 14.5260 |
| 4 | 697.6727 | 0.1103 | 13.1447 | 721.0998 | 0.1176 | 13.3514 |
| baseline 1: gaussian noise (steps=100) | 878.9229 | 0.1798 | 14.6848 | / | / | / |
| baseline 2: gaussian noise (steps=150) | 1147.3047 | 0.1040 | 11.9932 | / | / | / |
| baseline 3: random crop (scale: 0.7-0.9) | 613.9709 | 0.1349 | 14.2975 | / | / | / |

## 4. Results

4.1 Performance Evaluation of Participating Teams

This section presents a comprehensive analysis of all participating teams' performance during both the preliminary and final rounds, as detailed in the competition rankings. The leaderboard rankings were determined exclusively using the FVD score in the final round, which emphasizes semantic similarity evaluation. While image-based metrics (PSNR and SSIM) were recorded for reference, they were not considered in the ranking process due to challenges in ensuring fair weighting or simple averaging across these distinct evaluation criteria.

Table 4 presents the performance of the top four teams in the final round, along with baseline comparisons and their results in the preliminary round. As summarized, the top four solutions from the final round demonstrated competitive performance. Notably, Algorithm 1 achieved the best FVD score in both the preliminary and final rounds.

The baseline evaluations (baseline 1&2) were conducted by progressively introducing Gaussian noise to OCT images. In the baseline experimental setup, these noise-corrupted OCT images served as synthetic model predictions, while the original, unmodified OCT images were still maintained as ground truth references. All participating teams significantly outperformed these baselines in terms of video-based distance metrics. However, regarding image-based metrics, teams generally achieved lower SSIM and PSNR scores. This observation suggests that traditional pixel-level similarity metrics may be disproportionately affected by linear transformations, consequently limiting their effectiveness in capturing semantic relationships. To further investigate this phenomenon, we implemented an additional baseline (baseline 3) involving random cropping of 3D OCT images (cropping scale: 0.7-0.9). Despite the intentional information loss from cropping, the FVD scores remained relatively stable, while SSIM and PSNR performance deteriorated significantly. These results strongly suggest that for future OCT image generation research: (1) high-level semantic evaluation metrics should be prioritized, and (2) domain-specific adaptations of FVD may offer promising avenues for methodological advancement.

4.2 Influence of Generative Models in the Challenge

Generative models each present distinct advantages and limitations, and research teams proposed targeted improvements to optimize their performance for this application. Based on the model structures and 3D learning strategies outlined in Table 3, we examine the core enhancements developed by the research teams.

Algorithm1 adopted the LDM, a variant of diffusion models known for high generation fidelity, strong mode coverage, and sample diversity[29]. LDMs further reduce computational costs while preserving fine-grained details[19]. Given the limited dataset size in the challenge, Algorithm1 incorporated pretrained fundus foundation models to effectively integrate prior ophthalmic knowledge into the generative framework. Additionally, they leveraged supplementary datasets to pretrain the diffusion model, enhancing its ability to capture data diversity. To improve generalization, they introduced collaborative data augmentation, simultaneously processing fundus and 3D OCT images to strengthen cross-modal learning.

Both Algorithm 2 and 3 employed conditional GANs, addressing well-known challenges such as training instability and mode collapse, which can reduce sample diversity. Their solutions incorporated distinct enhancements. Algorithm 2 introduced quality control mechanisms within the generation pipeline and performed ROI cropping on fundus images to better align them with corresponding OCT B-scans. Algorithm 3 focused on architectural improvements, training multiple sub-models, each dedicated to generating OCT B-scans from different positional perspectives. This modular approach improved both robustness and generalization.

Algorithm 4 proposed an autoencoder-style model, benefiting from stable training dynamics but facing the typical drawback of blurrier outputs compared to GANs due to noise injection and imperfect reconstruction. To mitigate this, they incorporated SSIM loss to preserve structural similarity between generated and real images, and Smooth L1 loss[30] to enhance pixel-level accuracy and robustness against outliers. These refinements significantly improved the fidelity of the generated OCT images.

In summary, these solutions proposed in the APTOS-2024 challenge highlight that while diffusion models excel in fidelity and diversity, GAN-based approaches remain competitive with proper stabilization techniques. Meanwhile, autoencoders, although more stable, require careful design of the loss function to match the perceptual quality of adversarial methods. Future work may explore hybrid architectures or multimodal foundation model integration[31,32] to further advance OCT generation.

5. Discussion

5.1 Summary of the Challenge

This challenge established the first multimodal benchmark for the automatic generation of 3D OCT images from 2D fundus images, providing an objective framework for evaluating generation fidelity in fundus-to-OCT translation. It addresses a critical gap in ophthalmic imaging research and lays a robust foundation for future investigations in cross-modal medical image synthesis.

With 342 participating teams, 42 preliminary submissions, and 9 finalists, the competition fostered significant methodological advancements in translating 2D fundus photographs into 3D OCT data. Analysis of the top-performing approaches revealed that successful solutions universally incorporated following critical elements. Effective methodologies explored in the challenge combined optimized data preprocessing, paired fundus-OCT augmentation techniques, and tailored model architectures with corresponding 3D learning. Beyond these technical implementations, the leading approaches strategically leveraged pretrained foundation models and supplemental OCT datasets to enhance generation quality, demonstrating particular effectiveness in addressing the data scarcity limitations inherent in this challenging cross-modal translation task, as clearly evidenced by the superior performance of Algorithm 1 in the challenge evaluation.

The results demonstrate the technical feasibility of reconstructing 3D retinal structures from 2D images. This has important implications for enhancing diagnostic capabilities in settings where OCT acquisition is limited or unavailable, offering a pathway to more accessible and informative retinal imaging. Moreover, it provides a valuable reference for cross-modal generation in other medical imaging contexts.

The APTOS challenge series continues to drive progress in ophthalmic AI by curating datasets and benchmarks focused on clinical relevance. Past challenges have included diabetic retinopathy classification (2019), prediction of treatment response in diabetic macular edema using OCT (2021)[1], angiography report generation (2023)[33], and the current fundus-to-OCT generation (2024). With growing interest in generative AI, this series is poised to accelerate innovation in multimodal learning. The success of this year's leading approaches highlights the importance of combining architectural innovation with strategic use of auxiliary data and domain knowledge.

5.2 Limitations and future direction

Despite the dataset's considerable size and diversity, there remain opportunities for enhancement. Future iterations could involve data collection from a wider range of clinical sites and include a broader spectrum of ocular diseases. Increasing the heterogeneity of the dataset would help mitigate biases and enable more equitable model performance across both common and rare (long-tail) conditions.

A key limitation of the current dataset lies in the scope of the OCT data. The challenge focused primarily on partial OCT representations, rather than complete volumetric reconstructions. This restricts the clinical applicability of the generated OCT images, as comprehensive 3D volumes are often required to fully visualize retinal structures for accurate diagnosis and in-depth analysis.

Moreover, quantitative OCT metrics such as retinal layer thickness and macular curvature have substantial clinical relevance.[34,35] Including detailed anatomical annotations, such as retinal layer boundaries, would enrich the training data, enabling models to generate outputs that are not only visually realistic but also clinically meaningful. Such contour information would enhance the interpretability and diagnostic utility of the generated OCT images.

In addition to OCT, other imaging modalities like MRI offer complementary information, such as full-eye morphology.[36] Future challenges may consider incorporating MRI data to support more advanced multimodal modeling, potentially enabling a more holistic representation of ocular anatomy and further expanding the translational value of cross-modal generative approaches.

Firstly, the dataset can be further improved despite its size and diversity. For example, the data collection could involve more hospitals and cover a broader range of ocular diseases. This would increase the dataset's heterogeneity, enabling the model to perform more fairly for both common conditions and long-tail distributions. Secondly, the OCT images generated in the challenge dataset are not yet comprehensive enough, such as the generation of complete OCT volumes. This limits the application of this AI tool in clinical practice and research, as it may not capture the full 3D structure and details necessary for accurate diagnosis and analysis. Additionally, the retinal layer thickness and macular curvature obtained through OCT quantitative analysis have broad application prospects[34,35]. If the dataset includes contour information for the various retinal layers, it could add more value to the generative model's outputs, while also providing more detailed information for the model to learn from. This would enhance the model's ability to generate more accurate and clinically useful OCT images. Furthermore, except for OCT, MRI can provide a full eye shape of the globe,[36] future challenges may incorporate MRI to achieve more advanced multimodal modeling.

## 6. Conclusion

This challenge bridges critical gaps in fundus-to-3D-OCT translation while fostering unprecedented collaboration to address this clinically significant task. Despite the technical complexity of generating 3D OCT images from 2D fundus images, the competition has inspired exceptionally innovative solutions. The overwhelming participation and quality of submissions confirm the challenge's success in stimulating meaningful progress. Efforts have been dedicated to organizing a competition that is both challenging and fair, aimed at promoting collaboration and advancing knowledge within the AI and ophthalmology communities.

**Funding**

The study was supported by the Global STEM Professorship Scheme (P0046113) and Henry G. Leong Endowed Professorship in Elderly Vision Health.

**Declaration of competing interest**

The authors have no conflicts of interest to declare.

**Data availability**

The dataset is available to the public on the official website at https://asiateleophth.org/cross-country-datasets/. Interested applicants are required to fill out the application form on the APTOS website, and sign the Data Use Agreement. The application will undergo review, and APTOS will respond within 7 working days.

**Acknowledgments**

We thank the InnoHK HKSAR Government for providing valuable supports.

The research work described in this paper was conducted in the JC STEM Lab of Innovative Light Therapy for Eye Diseases funded by The Hong Kong Jockey Club Charities Trust.

**Appendix. Details of competing solutions**

In this appendix, we provide detailed descriptions of Algorithm 1-4. Information about the top four research teams is presented in Table A.1.

**Table A.1**

Information of the top four teams. From top to bottom, from first to fourth place. The name of the third-place team is in Chinese, but it is shown in this report in its translated English form.

| Algorithm | Team name | Affiliation |
|---|---|---|
| 1 | NJUST-EYE | Nanjing University of Science and Technology |
| 2 | MC1 | Mayo Clinic |
| 3 | Mickey Mouse Clubhouse | China Medical University |
| 4 | ViCBiC | University of Electronic Science and Technology of China |

A.1 Algorithm 1 (NJUST-EYE)

NJUST-EYE designed the model based on the latent diffusion model (LDM), incorporating a fundus encoder, a spatial map encoder, and several intra-batch self-attention blocks (Fig. A.1a). 3D OCT images are encoded by the image-level encoder for each OCT B scan, while the diffusion model is trained in the low-dimensional latent space. The final generated results are obtained through the image-level decoder. This process also involves the guidance of fundus image information and the spatial map. The denoising U-Net in the diffusion model consists of several residual blocks with self-attention. For the condition features extracted from the fundus images, interaction is achieved by introducing a cross-attention mechanism and intra-batch self-attention. The intra-batch self-attention block treats batch of images as a sequence and automatically learn the correlation between six OCT images (Fig. A.1b). In the fundus encoder, NJUST-EYE used three pretrained foundation models to encode color fundus images and extract disease-related features (Fig. A.1c). The spatial map encoder encodes the spatial map to the same size as the latent representations of the images (Fig. A.1d).

In addition to the dataset provided by the challenge, three additional OCT datasets were used. For the training data provided by the challenge, both the OCT and fundus images were cropped to regions of interest (ROIs) (Fig. A.2a-c). The data augmentation used was a specifically designed flip enhancement, which involves flipping the color fundus and the corresponding OCT images horizontally. The additional OCT datasets only involved cropping of the OCT images, as shown in Fig. A.2d-f.

For implement details, NJUST-EYE employed a two-stage training strategy. In the first stage, NJUST-EYE trained a standard LDM using 2D OCT images. After this phase, the model was capable of generating individual OCT images unconditionally. During this stage, NJUST-EYE utilized 4 RTX 4090 GPUs, with a batch size of 18 per GPU. The model was trained for 100 epochs, taking approximately 2 days to complete. In the second stage, NJUST-EYE trained a conditional OCT sequence (six images) generation model. NJUST-EYE integrated additional modules into the LDM and finetuned the parameters of the cross-attention blocks, intra-batch self-attention blocks, and the spatial map encoder. In this stage, NJUST-EYE used 1 RTX 4090 GPU with a batch size of 6. NJUST-EYE trained 10 epochs and spend about 30 minutes.

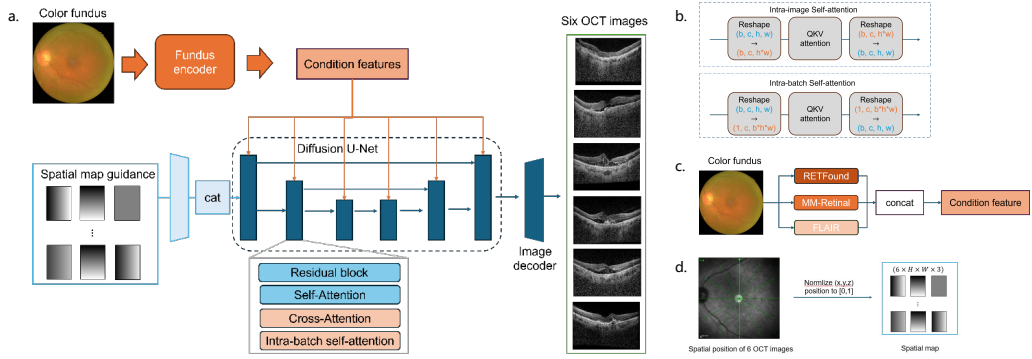

**Fig. A.1. Overview of the generative framework proposed by NJUST-EYE.** a) The model was designed based on LDM, incorporating a fundus encoder, a spatial map encoder, and several intra-batch self-attention blocks. b) Illustration and comparison of (intra-image) self-attention and intra-batch self-attention. c) Condition feature is extracted from fundus images from multiple foundation models. d) The spatial map encoder encodes the spatial map to the same size as the latent representations of the images.

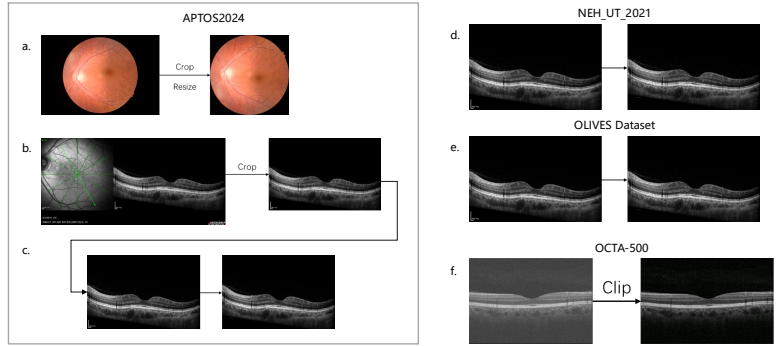

**Fig. A.2. Data pre-processing in NJUST-EYE.** a) Fundus images in APTOS2024: Crop out the excess black background of the color fundus and resize to 224*224. b) OCT image extraction. c) Remove the bottom left ruler in the OCT image and take the pixel value from the neighborhood to overwrite it. d) NEH_UT_2021: Remove the bottom left ruler in the OCT image and take the pixel value from the neighborhood to overwrite it. e) OLIVES: Remove the bottom left ruler in the OCT image and take the pixel value from the neighborhood to overwrite it. f) Style alignment, pixel value truncation (min=62 and max=255 when exporting data), and normalization.

A.2 Algorithm 2 (MC1)

In the method proposed by MC1, the Pix2Pix conditional generative adversarial network as originally described was used via the publicly available implementation as a starting point for subsequent customizations. A U-Net architecture was used for the generator and a $70 \times 70$ PatchGAN classifier for the discriminator. A vanilla GAN binary cross-entropy loss function as described in the original paper was used. For pre-processing and augmentation, each originally supplied training fundus photo was duplicated six times and rotated, aligned, and cropped to properly orient and focus on the central ROI as define by each of the six per patient supplied OCT B scans. For each fundus image, the vertical central 20 percent and the horizonal central 60 percent was retained and the rest of the image was masked using black bars. For the OCT images, the left overview and lower legend sections were discarded. These processed fundus and OCT B scans were then each resized to $768 \times 496$ pixels and combined into final paired training images (Fig. A.3).

Validation and final test inference were both performed in the cloud on the same training platform as described above. For the final round test session, inference was performed using the 32 "best" scoring preliminary round test models. In some models, for the worst 2-3% of the fundus images, a custom SSIM model was used to find and replace these poor-quality images with OCT images from the training set that are most similar. This replacement sometimes led to small improvements in the test results, but in other cases, it caused the performance to worsen. Essentially, the replacement had mixed effects—sometimes helping, sometimes hurting the final outcomes.

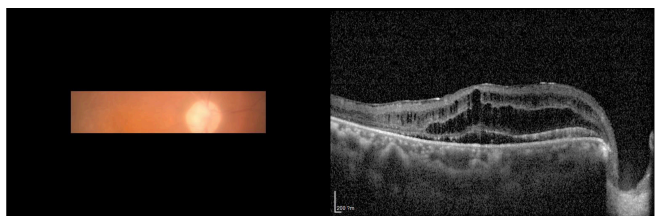

**Fig. A.3.** Example of post-processed masked, rotated, aligned, and resized training image for a single OCT slice.

Training and optimization were performed using a single GPU, most often the NVIDIA A6000, in the cloud using Digital Ocean Paperspace Gradient. Microsoft One and Digital Ocean Spaces were used for storage. Total cloud compute and storage costs were approximately $60 USD monthly. Typical training times were approximately 4-5 minute per epoch or 7-8 hours for a 100 epoch training of a unique model configuration. Hundreds of model parameter optimizations were attempted. The best performing final models shared the following parameter characteristics: epochs between 95 and 195, a 256 random crop size, lambda 100, learning rate between 0.0001 and 0.0002, and 128 layers for both the generator and discriminator. Model performance was evaluated for each epoch by initial sample

image review followed by on-device scoring using a custom Fréchet Inception Distance (FID) calculation based on the PyTorch metrics FID calculation using a sum of scores for 64, 192, and 768 features.

A.3 Algorithm 3 (Mickey Mouse Clubhouse)

The method proposed by Mickey Mouse Clubhouse was based on a noise injected conditional generative adversarial networks, Dynamic-Pix2Pix. Dynamic-Pix2Pix incorporates dynamic neural network theory, which effectively enhances the generalization capability of Pix2Pix on small sample datasets. Specifically, when the input is a fundus image, Dynamic-Pix2Pix focuses on learning the direct mapping relationship between the fundus image and the OCT image. When the input is a noise image, the generator of Dynamic-Pix2Pix concentrates on learning the distribution characteristics of the target OCT image. A U-Net architecture with an extra noise bottleneck module was used for the generator. The primary function of this module is to limit the transmission of noise information, ensuring that the generator does not simply treat noise as a feature of the input image. Positioned in the middle of the U-Net, the module reduces the number of channels for noise input and utilizes max-pooling layers to further filter out noise. This ensures that the generator can focus on the critical structures of the image rather than the noise itself. As a result, the generated OCT images appear smoother and more natural in terms of visual quality.

Due to the requirement of generating six different directional OCT B-scans for each fundus image, Mickey Mouse Clubhouse trained six independent sub-models. Before training, the fundus images and OCT B-scans were each cropped to the ROIs and resized to a uniform dimension of $256 \times 256$. The training was implemented on an NVIDIA RTX A6000 with a batch size of 8. The Adam optimizer was used, with an initial learning rate set to 0.0002. During the early stages of training (the first 100 epochs), Mickey Mouse Clubhouse kept the learning rate constant to ensure the network can stably learn the basic image features. In the later stages (the subsequent 100 epochs), the learning rate was adjusted using a cosine decay function, allowing it to smoothly decrease during the final phase of training. During the inference stage, the final prediction was obtained by combining the outputs of the six independent sub-models.

A.4 Algorithm 4 (ViCBiC)

The model architecture adopted by ViCBiC was designed based on the decomposition of fundus images. All fundus images and OCT images in the training set were first resized to $256 \times 256$. Data augmentation includes random contrast, brightness, saturation adjustment, gamma correction, Gaussian blur, and Gaussian noise to fundus images. Then, fundus images were extracted with corresponding six sequences ($6 \times 8 \times 256$, each sequence is an 8 pixel-width vector of 256 dimention) along the six OCT scanning directions. In the ViCBiC model, as illustrated in Fig. A.5, three layers of Multilayer Perceptrons (MLP) are used to process the input sequence, resulting in a feature map with the shape $1 \times 6 \times 256 \times 256$. The output of this MLP processing is then passed through a U-Net architecture, which employs a sigmoid activation function at the output layer to generate OCT images. The U-Net consists of five layers of symmetric encoder and decoder, each having 16 channels per layer. Within each layer of the U-Net, there are 3 Conv (3 × 3)-BatchNorm-ReLU blocks, which help capture hierarchical features and refine the generated images.

The model optimization process involves several key components aimed at improving the performance and accuracy of the generative model. Starting with an initial learning rate of 0.001, the AdamW optimizer is used with a weight decay of 5e-4 to ensure regularization and prevent overfitting. A batch size of 32 is chosen to allow efficient training, and the model is trained for 100 epochs to give it enough time to learn the complex mapping between decomposed fundus and OCT images. The learning rate is adjusted using the CosineAnnealingLR scheduler, gradually decreasing over time to facilitate smoother convergence as training progresses. For the loss function, both SSIM loss and Smooth L1 loss are employed. SSIM loss helps maintain structural similarity between the generated and target images, which is crucial for image quality, while Smooth L1 loss ensures pixel-level accuracy and robustness against outliers.

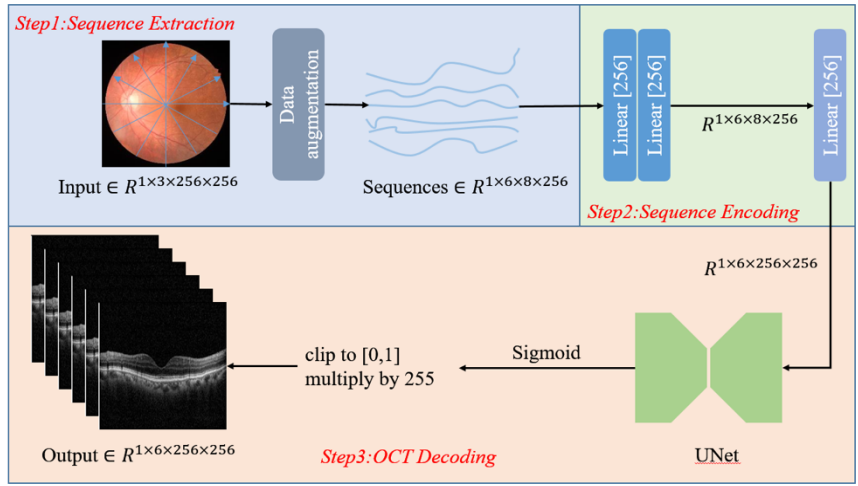

**Fig. A.5.** Model architecture designed by ViCBiC. An autoencoder-style generative framework for translating decomposed sequences of fundus images to OCT images.